\documentclass[11pt,a4paper]{article}
\usepackage[hyperref]{eacl2021}
\usepackage{times}
\usepackage{latexsym}

\usepackage{microtype}
\usepackage{booktabs}
\usepackage{multirow}
\usepackage{graphicx}
\usepackage{hyperref}
\usepackage{tabularx}
\usepackage{tikz}
\usepackage{listings}

\aclfinalcopy 

\makeatletter
\def\thanks#1{\protected@xdef\@thanks{\@thanks
        \protect\footnotetext{#1}}}
\makeatother

\title{Towards Modelling Coherence in Spoken Discourse}

\author{Rajaswa Patil\thanks{\hspace{2mm}$^{*}$Authors (listed in alphabetical order) contributed equally.}$^{*}$  \\
IIIT Delhi, BITS Pilani \\
\small \texttt{rajaswa.patil@midas.center} \And

Yaman Kumar Singla$^{*}$ \\
IIIT Delhi, Adobe, SUNY at Buffalo \\
\small \texttt{yamank@iiitd.ac.in} \AND

\hspace{-6mm} Rajiv Ratn Shah \\
\hspace{-6mm} IIIT Delhi \\
\small \texttt{rajivratn@iiitd.ac.in} \And

\hspace{-10mm} Mika Hama \\
\hspace{-10mm} Second Language Testing Inc.\\
\small \texttt{mika.hama@2lti.com} \And

Roger Zimmermann \\
National University of Singapore\\
\small \texttt{rogerz@comp.nus.edu.sg}

}

\date{}

\begin{document}
\maketitle
\begin{abstract} 
 While there has been significant progress towards modelling coherence in written discourse, the work in modelling spoken discourse coherence has been quite limited. Unlike the coherence in text, coherence in spoken discourse is also dependent on the prosodic and acoustic patterns in speech. In this paper, we model coherence in spoken discourse with audio-based coherence models. We perform experiments with four coherence-related tasks with spoken discourses. In our experiments, we evaluate machine-generated speech against the speech delivered by expert human speakers. We also compare the spoken discourses generated by human language learners of varying language proficiency levels. Our results show that incorporating the audio modality along with the text benefits the coherence models in performing downstream coherence related tasks with spoken discourses.
\end{abstract}

\section{Introduction}
Discourse coherence is a property of the organization and semantic relationships within a structured group of written or spoken utterances. Coherence can be used as an auxiliary metric to evaluate the quality of a given discourse. Previously, discourse coherence has been used to evaluate written discourses for tasks like essay scoring and readability assessment \citep{10.1017/S1351324903003206, burstein-etal-2010-using, mesgar-strube-2018-neural}. Coherence based metrics and objectives have also been used to evaluate and improve text-based artificial natural language generation systems \citep{10.5555/1642293.1642467, NIPS2015_5776, kiddon-etal-2016-globally} for tasks like text-summarization \citep{barzilay-lapata-2008-modeling, nenkova2011automatic}, machine-translation \citep{smith-etal-2016-trouble}, language modelling \citep{iter-etal-2020-pretraining, Lan2020ALBERT:} and conversation thread reconstruction \citep{joty-etal-2018-coherence}. Therefore, modelling discourse coherence has become an essential task in computational linguistics with a variety of downstream applications. 

Most of the previous work dealing with coherence modelling has been limited to text-based coherence \cite{joty-etal-2019-discourse}, and there has been limited work on modelling spoken discourse coherence as a task \citep{wang-etal-2013-coherence}. The few studies which have tried to work on spoken discourse have done so by transcribing speech and then applying text-coherence modelling methods on it \cite{wang-etal-2019-using}. Modelling coherence of a spoken discourse with its text-transcriptions is an inherently lossy and challenging task \citep{tappe-schilder-1998-coherence-spoken}. On the one hand, crucial cues of speech such as pauses, tonal variations, speed changes, stress, rhythm and intensity are lost while transcribing it, and on the other, the transcription is itself a cumbersome process with involved logistics and errors from automatic speech recognition (ASR) systems \citep{ERRATTAHI201832}. Various studies in linguistics have highlighted the importance of prosody in providing a structure to the spoken discourse \citep{10.5555/898272, degand2009identifying}. By incorporating the audio modality along with the text-transcription of the speech, we can better model the coherence in spoken discourse. 
In this paper, we make the following contributions:
\begin{itemize}
    \item We model coherence in spoken discourse with audio and text-based neural coherence models, which to the best of our knowledge, has not been previously explored.

    \item We perform experiments with four coherence-related tasks to inspect the role of audio modality in modelling spoken discourse coherence: \emph{Speaker Change Detection} (SCD), \emph{Artificial Speech Evaluation} (ASE), \emph{Topic Change Detection} (TCD) and \emph{Speech Response Scoring} (SRS).

    \item We show that incorporating the audio modality from speech benefits a coherence model in capturing discourse coherence and in performing various downstream applications with human and machine-generated speech.

    \item We also show that the audio modality becomes significantly important while modelling coherence in the less structured and noisy spoken discourses.
\end{itemize}

\section{Background}
Early work in modelling discourse coherence focused on extracting features based on the Centering Theory \citep{grosz-etal-1995-centering} and entity transitions in the text \citep{10.5555/1642293.1642467, elsner-etal-2007-unified}. \citet{barzilay-lapata-2008-modeling} introduced the entity grid representation of discourse, which was based on discourse entities and their grammatical role transitions. The entity grid model was further improved for coherence-related tasks by \citet{elsner-charniak-2011-extending},  \citet{feng-hirst-2012-extending} and \citet{louis-nenkova-2012-coherence}. Parallely, many works \citep{pitler-nenkova-2008-revisiting, lin-etal-2011-automatically, feng-etal-2014-impact} performed coherence-related tasks based on discourse relations in the text, parsed with theories like the Rhetorical Structure Theory (RST) \citep{thompson1987rhetorical} and the Lexicalized Tree Adjoining Grammar for discourse (D-LTAG)  \citep{WEBBER2004751} with the Penn Discourse Treebank (PDTB) \citep{prasad-etal-2008-penn} styled annotations. Notably, the features based on the RST-encodings were found to be useful for modelling coherence in spoken discourse \citep{wang-etal-2019-using} and more efficient than the PDTB-encodings for modelling text coherence as well \citep{feng-etal-2014-impact}. Further, \citet{guinaudeau-strube-2013-graph} and  \citet{mesgar-strube-2015-graph} proposed graph representation-based approaches to model coherence in text.

Following the advances in deep neural network architectures and distributed representations, there has been much progress towards developing neural models of discourse coherence which provide significant performance gains over the traditional feature-based models. The entity grid representation of discourse got extended with neural architectures. \citet{tien-nguyen-joty-2017-neural} proposed the neural entity grid model, which performed convolutions over the entity grid representations. Further, \citet{joty-etal-2018-coherence} lexicalized the neural entity grid model by attaching the entities to their respective grammatical roles in the entity grid embeddings.

Neural coherence models can be broadly classified into two categories: generative models and discriminative models. On the one hand, generative coherence models deal with modelling the conditional probabilities of a sentence being coherent with a given set of preceding sentences \citep{li-jurafsky-2017-neural, logeswaran2018sentence}. On the other hand, discriminative coherence models are trained to classify coherent and incoherent texts. It has been previously shown that modelling local coherence with discriminative models can be beneficial for capturing both the local and the global contexts of coherence with good approximation \citep{moon-etal-2019-unified, xu-etal-2019-cross}. Similarly, capturing relations and similarities between sentences at a local level with neural models can be helpful with coherence-related tasks \citep{li-hovy-2014-model, mesgar-strube-2018-neural}. Recent work in coherence modelling has focused on building models in open-domain \citep{li-jurafsky-2017-neural} and cross-domain \citep{xu-etal-2019-cross} settings. More recently \citet{lai-tetreault-2018-discourse} built coherence models and datasets for real-world texts. Coherence modelling has also been beneficial for the recent breakthroughs in transformer-based language modelling \citep{iter-etal-2020-pretraining, Lan2020ALBERT:}. Some recent work has also focused on building benchmarks for applying coherence models in evaluating text-based natural language generation systems \citep{mohiuddin2020coheval}.

Coherence deals with the \textit{perception} of the discourse rather than the discourse content itself \citep{COHEN200173, wang2014short, li2017discourse}. While the perception of a written discourse is only affected by the semantic organization of its lexical content, the perception of a spoken discourse is additionally dependent on its prosodic and acoustic features \citep{hirschberg-nakatani-1996-prosodic}. Previous work in linguistics has highlighted the role of prosody in defining the structure for spoken discourse. \citet{10.5555/898272} performed experiments with cooperative dialogues and demonstrated the role of prosodic information in defining the topic structure of a given spoken discourse. Further, \citet{degand2009identifying} introduced prosodic segmentation to define basic discourse units in speech. 

Various previous studies have used prosodic attributes to perform coherence-related tasks with spoken discourse. \citet{10.5555/898272} analysed the role of intonation and pause durations in modelling semantic relationships between discourse utterances at topic boundaries. Further, \citet{doi:10.1162/089120101300346796} used duration and pitch based features to perform the task of topic segmentation, which is closely related to both the local and global coherence of spoken discourse. \citet{stifelman1995discourse} used pitch patterns to perform emphasis detection with automated discourse segmentation. This was further used to summarize and skim through spoken discourses, a task which is highly relevant to the comprehensibility and the perception of spoken discourse.

Apart from the above mentioned tasks, automated speech scoring is an another important application of modelling discourse structure and coherence. Explicitly annotated coherence based measures \citep{wang-etal-2013-coherence, Wang2017} and features extracted from discourse structures in text-transcriptions of spoken discourses \citep{wang-etal-2019-using} help in improving the performance of the automated speech scoring systems significantly. Unlike the work done with essay scoring as an auxiliary evaluation task for coherence modelling, work in speech scoring has been limited to include discourse coherence related features from the text-transcriptions of spoken discourse along with other features relevant to speech scoring. 

\section{Methodology}
\subsection{Coherence Models for Spoken Discourse}
A coherence model for spoken discourse should be able to capture both the prosodic (pitch, intonation and stress) and the acoustic features (fundamental frequency, intensity and duration) of an audio sample. It is relatively easier to procure text datasets of varying levels of structure from sources like Wikipedia, social networks, and other online medium. Given the lack of such structured discourse-rich datasets for speech, the model should generalize beyond closed-domain settings \citep{li-jurafsky-2017-neural, xu-etal-2019-cross} and perform well on more open and cross-domain settings with limited training data \citep{iter-etal-2020-pretraining, mohiuddin2020coheval}. This becomes more important with spoken discourse as it has been shown that the audio modality is more vulnerable to change in data domain and background as compared to text \citep{yan2020handbook}. Given the above mentioned challenges and pitfalls related to modelling coherence in spoken discourse, an ideal audio-based coherence model should:

\begin{enumerate}
    \item learn a discourse coherence signal with a limited number of training samples
    
    \item generalize across speech samples which vary in terms of the acoustic and prosodic features of the speech audio (Ex: accent, gender, rhythm, age, speed and intonation) and differences in the background of data like recording quality, sampling frequency, background noise \textit{etc.}

    \item and, similar to the text domain, generalize across speech samples which vary in terms of the spoken discourse's topic and content. 

\end{enumerate}

\begin{figure}[thbp]
  \center
  \includegraphics[width=0.60\columnwidth]{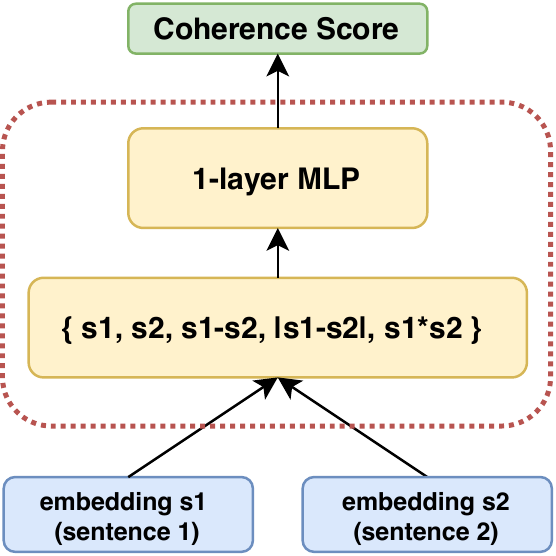}
  \caption{The local discrimination coherence model.}
  \label{model-fig}
\end{figure}

The local discrimination algorithm proposed by \citet{xu-etal-2019-cross} is designed to maximize the local coherence scores of adjacent (positive) pairs of sentences and minimize the local coherence scores for the non-adjacent (negative) pairs of sentences in a given discourse. Unlike the older discrimination models which suffer with class-imbalance between the coherent and incoherent permutations of written discourses, this approach captures local coherence with an effective negative sampling of the incoherent non-adjacent sentences.  The model takes in a pair of sentence representations as an input, which are further passed through a multi-layered perceptron with a single hidden layer (Figure~\ref{model-fig}) to obtain a local coherence score. Experiments done by \citet{xu-etal-2019-cross} show that the global aspects of coherence can be  approximated by using the local coherence scores from their models with techniques like score-averaging across the discourse. Further, the local discrimination model learns to generalize in open-domain as well as cross-domain settings (as shown by their sentence-ordering and paragraph-reconstruction experiments with domain-separated Wikipedia articles), and is agnostic to the modality and background of the input data. Hence, for all our experiments, we use the local discrimination model proposed by \citet{xu-etal-2019-cross}. 

\begin{figure*}[htbp]
  \center
  \includegraphics[width=1.0\textwidth]{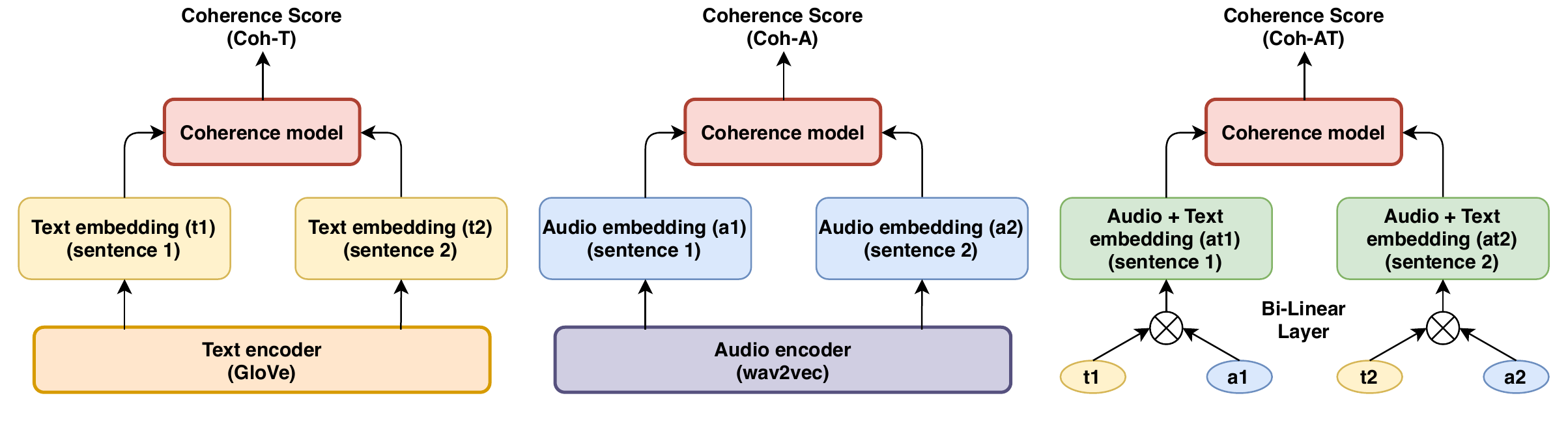}
  \caption{Model architecture across the three input settings: Coh-T, Coh-A and Coh-AT.}
  \label{architecture-fig}
\end{figure*}

To incorporate the audio modality into the coherence model, we encode an audio based sentence embedding, similar to the sentence embedding obtained from the text modality. We use a pre-trained audio language model: wav2vec \citep{Schneider_2019} to encode the audio segment of a sentence into its corresponding audio embedding as shown in Figure~\ref{architecture-fig}. Wav2vec is pre-trained with an unsupervised objective of next time-step prediction task for audio segments. This objective function aligns significantly with that of the local discrimination coherence model, providing rich audio representations for our task. Similarly, we use pre-trained GloVe embeddings \citep{pennington2014glove} to encode the text from the sentence into its corresponding text embedding. 

In order to inspect the role of the audio modality in modelling coherence for spoken discourses, we experiment with three different learning settings (Coh-T, Coh-A and Coh-AT) for the local discrimination coherence model (Figure~\ref{architecture-fig}). Similar to all the previous work, with the first setting, we just use the text modality as the input to the coherence model (\textbf{Coh-T}). In the second setting, we use only the audio modality as the input to the coherence model, to establish an audio-only control setting for our experiments (\textbf{Coh-A}). Finally, in the third setting we obtain a multimodal input representation by fusing the text and audio modalities together (\textbf{Coh-AT}). In order to get a minimal trainable aggregated fusion of the two modalities, we pass the audio and the text embeddings through a bi-linear layer as shown in Figure~\ref{architecture-fig}. Following \citet{xu-etal-2019-cross}'s approach, we aggregate the coherence scores from both the forward model \emph{(sentence-1, sentence-2)} and the backward model \emph{(sentence-2, sentence-1)}.

\subsection{Datasets}
We perform all our experiments with the debate speech samples from the IBM Debater dataset\footnote{\url{https://www.research.ibm.com/haifa/dept/vst/debating_data.shtml}} \citep{lavee2019towards}. The dataset consists of recordings of debate speeches delivered by nine expert debaters, with debate speeches on 200 distinct Wikipedia topics (such as social media, nuclear weapons, gambling, \textit{etc.}) as the debate motions. Each motion topic is contested with two debate recordings from distinct experts, resulting in a total number of 400 unique speech samples. This makes the dataset rich with a variety of coherent speeches spread across a variety of open-domain topics, providing a high quality training signal to our coherence models. We generate new datasets\footnote{The datasets are available \href{https://drive.google.com/drive/folders/1DPBC9RM5_zfNoCc1OKpd7jLl1P7b7_1y?usp=sharing}{here}} for various evaluation tasks (explained in Section~\ref{task:srs}) using the responses from the IBM Debater dataset. For this, we use the text-transcriptions from the debate speech recordings to synthesize artificial speech responses with a standard text-to-speech (TTS) system based on the Microsoft Speech API (SAPI5) \cite{sapi5}. We use two distinct TTS voices across all our experiments: \textbf{S1} (US-male voice) and \textbf{S2} (US-female voice). 

\begin{table}[ht]
\center
\begin{tabular}{@{}cccc@{}}
\toprule
\multirow{2}{*}{\textbf{Task}} & \multicolumn{3}{c}{\textbf{Number of speech responses}}  \\ \cmidrule(l){2-4} 
                               & \textbf{Train} & \textbf{Validation} & \textbf{Test}  \\ \midrule
\textbf{SCD\textsuperscript{*}}                   & -              & -                   & 398           \\
\textbf{ASE\textsuperscript{*}}                   & 197            & 78                  & 120           \\
\textbf{TCD\textsuperscript{*}}                   & -              & -                   & 786           \\ \midrule
\textbf{SRS}                   & 463            & 181                 & 234            \\ \bottomrule
\end{tabular}
\caption{Number of unique speech responses across the datasets for various evaluation tasks (*each unique speech response is further sampled with two TTS settings, resulting in twice as many evaluation samples).}
\label{tab:sample-stats}
\end{table}

Further, we also use another dataset of speech responses from non-native English language learners for one of the evaluation tasks. We use the speech samples delivered by human speakers to train the coherence models (task ASE and SRS), whereas the speech samples synthesized with the TTS voices are used for evaluating the models on various coherence-related tasks (task SCD, TCD, and ASE). Further details about the training and testing of coherence models on the evaluation tasks are given in Section~\ref{task:srs}. The statistics for all the datasets are mentioned in Table~\ref{tab:sample-stats}.

\subsection{Evaluation Tasks}
Empirical results from previous work in coherence modelling for text has shown that the traditional synthetic tasks like sentence ordering do not effectively capture the models' ability to perform downstream discourse coherence related tasks with real-world data \citep{lai-tetreault-2018-discourse, mohiuddin2020coheval}. An ideal coherence model should perform well (assign appropriate coherence scores) for spoken discourses delivered by humans as well as machine-generated speech. Keeping this in mind, we design four tasks for evaluating coherence in spoken discourse: \emph{Speaker Change Detection} (\textbf{SCD}), \emph{Artificial Speech Evaluation} (\textbf{ASE}), \emph{Discourse Topic Change Detection} (\textbf{TCD}) and \emph{Speech Response Scoring} (\textbf{SRS}). While the first three tasks focus on artificially generated speech and speech delivered by expert human speakers, the fourth task focuses on comparing the spoken discourses delivered by non-native language learners of varying proficiency levels.

\paragraph{Speaker Change Detection (SCD):}
Modelling coherence in a conversational setting is a very important task with various downstream applications \citep{joty-etal-2018-coherence, 10.1007/978-3-030-00671-6_37}. To obtain a structured conversational text-transcript from a given speech audio, we first need to perform speaker diarization. Hence, speaker diarization is an important aspect of modelling  conversational spoken discourse \citep{moattar2012review}. Acoustic cues play an important role in speaker segmentation of conversational speech \citep{Park_2018}. Since all the samples from the IBM Debater dataset \citep{lavee2019towards} are monologue debate speeches, we construct a new dataset for this task using these monologues. We sample a ten-sentences long segment from the middle of every response. The first five sentences from the sampled segment are synthesized with a TTS voice (S1) and the next five sentences are synthesized with another TTS voice (S2). Consequently, the overall synthesized speech response consists of a speaker change at the end of the fifth sentence, while maintaining a continuation in the discourse topic. This speaker change in the response can be detected with an audio-based coherence model, where the event of speaker change can be depicted with the least inter-sentence local coherence score. Further in a separate experiment, we reverse the order of speakers, such that the first five sentences are synthesized with TTS voice S2 and the next five sentences are synthesized with TTS voice S1.

\paragraph{Artificial Speech Evaluation (ASE):} 
Following the work done in evaluating the quality of text-based natural language generation systems with coherence-based measures \citep{10.5555/1642293.1642467, NIPS2015_5776, kiddon-etal-2016-globally}, we propose evaluating the TTS systems with coherence models of spoken discourse. TTS systems often face issues while naturalizing the synthesized speech, and continuous efforts are being made to make TTS-speech more human-like and intelligible across longer contexts \citep{taylor2009text}. Modelling discourse relations and coherence can benefit a TTS system in delivering expressive and intelligible speech \citep{delmonte-tripodi-2015-semantics}. In this task, we evaluate and compare a TTS speech sample which lacks a certain amount of prosodic variation in terms of intonation, stress and speaking rate, to the speech from a expert human speaker. The underlying hypothesis for this particular task is that the lack of prosodic variation in the TTS sample makes it relatively incoherent as compared to the human speech, even though the delivered lexical content is same across both the samples. Given that we train the coherence models with human speech only, a direct comparison between the coherence scores of human and machine generated speech might include some biases emerging from audio-data background. To handle this, we perform the comparison using the relative difference between the mean coherence scores of the positive and negative sentence pairs of a given speech response (explained further in Section~\ref{sec-ase-results}).

\paragraph{Topic Change Detection (TCD):} 
 Following the work done by \citet{doi:10.1162/089120101300346796} in automatic topic segmentation with prosodic cues, we propose a coherence-based topic change detection task for spoken discourse. Since, all the samples from the IBM Debater dataset are restricted to a single topic (debate motion), we construct a new dataset for this task using the responses from the IBM Debater dataset \citep{lavee2019towards}. These samples majorly evaluate the extent to which a model captures the prosodic features at topic boundaries \citep{10.5555/898272}. For constructing a sample, we select a five-sentences long segment from the middle of every response, so that the sampled segment represents a developed topic rather than introductory definitions or concluding statements. Subsequently, we combine it with a similarly sampled segment from a different motion topic. This results in a new ten-sentences long speech response which covers a particular topic in its first five sentences and a different topic in the next five sentences. We use a text-to-speech system to synthesize the speech audio for this newly generated response. The underlying hypothesis for this particular task is that the local inter-sentence coherence score should be the lowest for the fifth and the sixth sentence, depicting a change in discourse topic for the given speech response.

\paragraph{Speech Response Scoring (SRS):}
\label{task:srs}
Coherence scores follow a monotonic relationship with the holistic language proficiency grades. Previous work in text-based coherence modelling has used essay scoring as an auxiliary evaluation task \cite{10.1017/S1351324903003206, mesgar-strube-2018-neural}. Conversely, modelling coherence has also proved to be beneficial in essay-scoring benchmarks \citep{tay2017skipflow, li2018coherence}. In a like manner, coherence-based features extracted from the text-transcriptions of spoken discourses have proved to be useful in scoring speech responses from language learners and non-native English speakers \citep{wang-etal-2013-coherence, Wang2017}. Following this, we test our coherence models on a dataset of spoken discourses delivered by non-native English language learners from Philippines \citep{grover2020multimodal}. The dataset comprises of speech responses recorded in a test environment where the candidates are asked to respond to one of the $48$ distinct prompts. They are subsequently double scored by expert annotators using a holistic language proficiency level on a $6$-point CEFR scale \citeyearpar{council2001common}. Since there is no publicly available speech dataset with explicit coherence-score annotations, these proficiency grades provide us with an approximate human evaluation of coherence. Using this data, we construct pairs of speech responses, such that every pair contains speech responses from two different speakers, for the same prompt (to eliminate prompt-based biases in the holistic proficiency grading). Given such a pair of speech responses, we hypothesize that the response graded with higher holistic proficiency level, should be assigned higher coherence scores by a coherence model. 

Spoken discourse from non-native speakers is usually less structured as compared to a discourse delivered by a native expert speaker \citep{yan2020handbook}, making it more challenging to model discourse coherence for non-native speakers. The text-transcripts of the speech responses in the dataset are not structured with proper punctuation, which is needed to obtain sentence-level segments of the speech response. So, we punctuate the text-transcripts from the dataset with a punctuator model \citep{tilk2016}. Given the background and the pre-processing involved, this dataset is more noisy and challenging as compared to the IBM Debater dataset. Moreover, while the discourses from the IBM Debater dataset are more argumentative and informative in nature, the responses in this dataset are more descriptive and narrative in nature. Hence, both the datasets vastly differ in terms of discourse modes \citep{song-etal-2017-discourse, dhanwal-etal-2020-annotated}. 

\section{Experimental Setup}
\paragraph{Audio Processing:}
Across all our experiments, we use a speaking-rate of $150$ words per minute to synthesize speech responses with the TTS systems, a value which is recognized as the average speaking rate for a native US-English speaking adult \cite{ncvs2020}. Unlike structured texts, there are no explicit cues to perform a sentence-level segmentation in speech. We use a pre-trained Montreal Forced Aligner (MFA\footnote{\url{https://montreal-forced-aligner.readthedocs.io/}}) model \citep{mcauliffe2017montreal} and the punctuation from the structured text-transcriptions to get the sentence-level alignments for the speech audio files. Further, all the speech responses are resampled to a $16$kHz mono-channel audio file as required by the pre-trained wav2vec model.

\paragraph{Coherence Modelling:}
Following \citet{xu-etal-2019-cross}'s training protocol, we sample the incoherent pair of sentences within the same speech response. This avoids the model pitfalls related to the topic and speech based features with the local discrimination setting . Further, to make the model generalize well across various domains and sources of the data, we do not fine-tune the pre-trained audio and text encoders on the training data. We train the model to optimize the local coherence scoring based margin loss objective as shown in Equation~\ref{eq:loss-fn}, where $f\textsuperscript{+}$ and $f\textsuperscript{-}$ are coherence scores for the adjacent (coherent) and non-adjacent (incoherent) pairs of sentences, respectively. We use $50\%$ of the topics in the dataset to train our model, while the rest $20\%$ and $30\%$ of the topics are used for validation and testing purposes, respectively. The model parameters are optimized with Adam optimizer \citep{DBLP:journals/corr/KingmaB14} with a learning rate of $0.001$. We validate the models with an early stop callback on the validation loss, with a patience of two epochs. Given, the cross-domain adaptation abilities of the local discrimination model, we borrow the hyperparameter settings from \citet{xu-etal-2019-cross} and do not perform any extensive hyperparameter tuning during our experiments.

\begin{equation}
\label{eq:loss-fn}
    L(f\textsuperscript{+}, f\textsuperscript{-}) = \max(0,  5-f\textsuperscript{+}+f\textsuperscript{-})
\end{equation}

\begin{equation}
\label{eq:sentence}
    k_{change} = \textstyle \min_{k \in [1,N-1]} \{f\textsuperscript{+}_k\}
\end{equation}

\begin{equation}
\label{eq:response}
    coherence\ score = \frac{1}{N-1}\sum_{k=1}^{N-1} f\textsuperscript{+}_k
\end{equation}

While the SCD task and TCD task are evaluated at a local level with inter-sentence coherence scores (Equation~\ref{eq:sentence}), the SRS task is evaluated with response-level coherence scores as shown in Equation~\ref{eq:response}, where $N$ is the number of sentences in the speech response.

\section{Results and Discussion}
\paragraph{Task SCD:} 
For a ten-sentences long response, task SCD results in a nine-way classification setting. The top-k\footnote{here a prediction is considered to be correct if the ground-truth value lies in the first $k$ predictions given by the model} (\textit{k=1,2,3}) accuracy scores for the SCD task are shown in Table~\ref{tab:results-scd}. As expected, the Coh-T model fails to capture the speaker change boundaries (with the accuracy scores being almost equal to that of random guessing) due to the lack of access to the acoustic information from the speech audio. The Coh-A model shows impressive accuracy for this task, consistent across both the orders of speaker-change. The model predicts almost all the speaker change boundaries for \textit{k=3}. The Coh-AT model does not match up in performance against the Coh-A model, suggesting a difference in audio-based learning between the two input settings.

\begin{table}[ht]
\center
\begin{tabular}{@{}ccccc@{}}
\toprule
\textbf{Model}                   & \textbf{Change}              & \textbf{k=1} & \textbf{k=2} & \textbf{k=3} \\ \midrule
\textbf{Coh-T}                   & -                            & 0.0944             & 0.2041             & 0.3138             \\ \midrule
\multirow{2}{*}{\textbf{Coh-A}}  & \textbf{S1 $\rightarrow$ S2} & 0.9770             & 0.9898             & \textbf{0.9949}             \\
                                 & \textbf{S2 $\rightarrow$ S1} & 0.9796             & 0.9974             & \textbf{1.0000}             \\ \midrule
\multirow{2}{*}{\textbf{Coh-AT}} & \textbf{S1 $\rightarrow$ S2} & 0.7398             & 0.8954             & 0.9311             \\
                                 & \textbf{S2 $\rightarrow$ S1} & 0.5026             & 0.6913             & 0.7730             \\ \bottomrule 
\end{tabular}
\caption{Top-k accuracy scores for the SCD task.}
\label{tab:results-scd}
\end{table}

Negative sampling within the same speech response restricts the model to look at audio segments from different speakers under the audio-based settings. Consequently, the model is only exposed to small prosodic and acoustic variations from the same speaker during training. This shows that the local discrimination model captures even large acoustic changes in a given conversational spoken discourse, by modelling local coherence with cues from small acoustic and prosodic variations in monologue speech.

\begin{table}[ht]
\center
\resizebox{\columnwidth}{!}{%
\begin{tabular}{@{}ccccc@{}}
\toprule
\multicolumn{2}{c}{\textbf{Speaker}}                                                                           & \textbf{Coh-T} & \textbf{Coh-A} & \textbf{Coh-AT} \\ \midrule
\multirow{3}{*}{\textbf{\begin{tabular}[c]{@{}c@{}}Human \\ Expert\end{tabular}}}  & $f\textsuperscript{+}$    & 0.75           & 0.97           & 0.84            \\
                                                                                   & $f\textsuperscript{-}$    & -1.14          & -1.76          & -2.52           \\ \cmidrule(l){2-5} 
                                                                                   & \textbf{\% diff}          & -252\%         & -281.40\%      & \textbf{-400\%} \\ \midrule
\multirow{3}{*}{\textbf{\begin{tabular}[c]{@{}c@{}}TTS voice\\ (S1)\end{tabular}}} & $f\textsuperscript{+}$    & 0.75           & 1.76           & 1.08            \\
                                                                                   & $f\textsuperscript{-}$    & -1.14          & 1.62           & -0.92           \\ \cmidrule(l){2-5} 
                                                                                   & \textbf{\% diff}          & -252\%         & -7.90\%        & -185.20\%       \\ \midrule
\multirow{3}{*}{\textbf{\begin{tabular}[c]{@{}c@{}}TTS voice\\ (S2)\end{tabular}}} & $f\textsuperscript{+}$    & 0.75           & 2.73           & 0.77            \\
                                                                                   & $f\textsuperscript{-}$    & -1.14          & 2.55           & -1.39           \\ \cmidrule(l){2-5} 
                                                                                   & \textbf{\% diff}          & -252\%         & -6.60\%        & -280.50\%       \\ \midrule
\end{tabular}%
}
\caption{Mean coherence scores of positive (adjacent) and negative (non-adjacent) pairs of sentences from the speech samples in the test set for task ASE, along with the relative difference (\% diff) between them. A higher \% diff value is indicative of better coherence models and more coherent spoken discourses.}
\label{tab:results-ase}
\end{table}

\paragraph{Task ASE:} 
\label{sec-ase-results}
Given the difference in the audio data backgrounds for expert human speakers and TTS systems, we compare their coherence by monitoring the relative difference between the mean coherence scores of the coherent (adjacent) and incoherent (non-adjacent) pairs of sentences sampled from the speech responses generated by them (Table~\ref{tab:results-ase}). In accordance with the training objective function, the incoherent sentences are scored lesser than the coherent sentences (negative relative difference) across all the speaker and model settings. A higher relative difference between the coherence scores of coherent and incoherent pairs of sentences not only indicates the coherence model's ability to effectively model coherence (horizontal traversal across Table~\ref{tab:results-ase}), but it also indicates the speaker's ability to produce more coherent discourses (vertical traversal across Table~\ref{tab:results-ase}). While the Coh-T model gives a relative difference of $-252\%$ on the samples from human experts, the audio-based Coh-A and Coh-AT models give much higher relative differences of $-281.40\%$ and $-400\%$, respectively. This shows that incorporating the audio modality highly benefits a coherence model to capture the difference between coherent and incoherent samples. Further, while the Coh-T model is independent of any changes in the speech audio (same mean coherence scores of $0.75$ and $-1.14$ for all the speakers), comparing the relative differences across the speakers, we observe that the TTS voices S1 and S2 show significantly lower relative differences across both the audio-based settings ($-7.90\%$ and $-6.60\%$ for Coh-A and, $-185.20\%$ and $-280.50\%$ for Coh-AT, respectively). Hence, under the ASE task, we find that the speech synthesized by TTS systems is relatively incoherent and more difficult to perceive as compared to human-generated speech.

\begin{table}[ht]
\center
\begin{tabular}{@{}ccccc@{}}
\toprule
\textbf{Model}                   & \textbf{Speaker} & \textbf{k=1}          & \textbf{k=2}          & \textbf{k=3} \\ \midrule
\textbf{Coh-T}                   & -                & \textbf{0.2513}       & 0.3980                & 0.5434       \\ \midrule
\multirow{2}{*}{\textbf{Coh-A}}  & \textbf{S1}      & 0.1148                & 0.2156                & 0.3444       \\
                                 & \textbf{S2}      & 0.1135                & 0.2742                & 0.3801       \\ \midrule
\multirow{2}{*}{\textbf{Coh-AT}} & \textbf{S1}      & 0.2385                & \textbf{0.4031}       & 0.5383       \\
                                 & \textbf{S2}      & 0.2449                & \textbf{0.4056}       & \textbf{0.5663}       \\ \bottomrule 
\end{tabular}
\caption{Top-k accuracy scores for the TCD task.}
\label{tab:results-tcd}
\end{table}

\paragraph{Task TCD:} 
Similar to the SCD task, the TCD task comes up with a nine-way classification setting. The top-k (\textit{k=1,2,3}) accuracy scores for the TCD task are shown in Table~\ref{tab:results-tcd}. The Coh-A model does not perform well on the TCD task individually, with the accuracy scores being almost equal to that of random guessing. While the Coh-T model slightly outperforms the Coh-AT model for \textit{k=1}, the Coh-AT model shows slight improvements over the Coh-T model for \textit{k=2} with both the TTS voices S1 and S2. Further, for \textit{k=3}, the Coh-AT model shows significant improvement over the Coh-T model for TTS voice S2. Even though topic segmentation is predominantly a text-based task, the slight improvements shown by Coh-AT model over the text-only settings can be explained by the presence of cues related to the prosodic patterns observed at topic boundaries \citep{10.5555/898272}.

\begin{table}[ht]
\center
\begin{tabular}{@{}cccc@{}}
\toprule
               & \textbf{Coh-T} & \textbf{Coh-A}  & \textbf{Coh-AT} \\ \midrule
\textbf{Accuracy}  & 0.4641         & \textbf{0.7046} & 0.5569          \\ \bottomrule
\end{tabular}
\caption{Accuracy scores for the SRS task with non-native speech dataset.}
\label{tab:results-srs}
\end{table}

\paragraph{Task SRS:} 
For this task, we monitor the accuracy scores for the binary classification setting based on the holistic language proficiency grades, using response-level coherence scores as the proficiency measure (Table~\ref{tab:results-srs}). While the Coh-A model performs significantly well with an accuracy score of $0.70$ on the test set, the Coh-T and Coh-AT models fail to perform well. Given the lack of structure in non-native speech and the noise in the text-transcriptions of the speech-responses, text-based settings do not capture the complex holistic grades efficiently. On the other hand, the audio-based setting seems to be resistant to this lack of structure and noise in transcriptions and it effectively captures the holistic language proficiency grades.

The near-perfect performance of the Coh-A model in predicting the speaker changes suggests that modelling coherence with the audio modality can turn out to be quite beneficial for a variety of discourse related tasks in conversational speech such as speech act detection, conversation disentanglement, etc. To further the efforts made in naturalizing the speech synthesized with text-to-speech systems, one can come up with better coherence-based objectives to train the TTS systems. Building up on the topic change detection task, coherence models for spoken discourses can be also evaluated on related downstream applications like topic-segmentation in spoken lectures, podcasts, political spoken discourses, etc. Further, the significantly higher performance of the Coh-A model on the SRS task shows that modelling coherence with audio modality can highly compensate for the lack of structure and errors in text-transcriptions of the speech. This can be quite useful while modelling coherence with data from non-native speakers, language learners or while using error-prone text-transcriptions from automatic speech recognition (ASR) systems.

\section{Conclusion}
In this paper, we performed experiments with four coherence-related tasks for spoken discourse. In our experiments, we compare the speech synthesized with text-to-speech systems against the expert human speakers. We also evaluate coherence in spoken discourses delivered by non-native language learners of varying language proficiency levels. Our experiments show that incorporating the audio-modality betters the coherence-modelling for spoken discourses significantly.

\bibliography{anthology,eacl2021}
\bibliographystyle{acl_natbib}

\end{document}